\documentclass[letterpaper]{article}
\usepackage{aaai2026} 
\usepackage{times} 
\usepackage{helvet} 
\usepackage{courier} 
\usepackage[hyphens]{url} 
\usepackage{graphicx} 
\usepackage{natbib} 
\usepackage{caption} 
\usepackage{algorithm}
\usepackage{algorithmic}

\usepackage{newfloat}
\usepackage{listings}
\DeclareCaptionStyle{ruled}{labelfont=normalfont,labelsep=colon,strut=off} 
\floatstyle{ruled}
\newfloat{listing}{tb}{lst}{}
\floatname{listing}{Listing}
\usepackage{amsmath}
\usepackage{booktabs}
\usepackage{amssymb}
\usepackage{multirow}
\usepackage{makecell}
\usepackage{pifont}

\title{MambaFlow: A Mamba-Centric Architecture for End-to-End Optical Flow Estimation}
\author{
    Juntian Du\textsuperscript{\rm 1},
    Zhihu Zhou\textsuperscript{\rm 1},
    Runzhe Zhang\textsuperscript{\rm 1},
    Yuan Sun\textsuperscript{\rm 1},
    Pinyi Chen\textsuperscript{\rm 1},
    Keji Mao\textsuperscript{\rm 1}
}
\affiliations{
    \textsuperscript{\rm 1}Zhejiang University of Technology\\
    djt@zjut.edu.cn,
    zhouzhihu@zjut.edu.cn,
    mupamupa0.0@gmail.com,
    221124120198@zjut.edu.cn,
    chenpinyi@zjut.edu.cn,
    maokeji@zjut.edu.cn
}

\usepackage{bibentry}

\nocopyright

\begin{document}

\maketitle

\begin{abstract}
Recently, the Mamba architecture has demonstrated significant successes in various computer vision tasks, such as classification and segmentation. However, its application to optical flow estimation remains unexplored. In this paper, we introduce MambaFlow, a novel framework designed to leverage the high accuracy and efficiency of the Mamba architecture for capturing locally correlated features while preserving global information in end-to-end optical flow estimation. To our knowledge, MambaFlow is the first architecture centered around the Mamba design tailored specifically for optical flow estimation. It comprises two key components: (1) PolyMamba, which enhances feature representation through a dual-Mamba architecture, incorporating a Self-Mamba module for intra-token modeling and a Cross-Mamba module for inter-modality interaction, enabling both deep contextualization and effective feature fusion; and (2) PulseMamba, which leverages an Attention Guidance Aggregator (AGA) to adaptively integrate features with dynamically learned weights in contrast to naive concatenation, and then employs the intrinsic recurrent mechanism of Mamba to perform autoregressive flow decoding, facilitating efficient flow information dissemination. Extensive experiments demonstrate that MambaFlow achieves remarkable results comparable to mainstream methods on benchmark datasets. Compared to SEA-RAFT, MambaFlow attains higher accuracy on the Sintel benchmark, demonstrating stronger potential for real-world deployment on resource-constrained devices. The source code will be made publicly available upon acceptance of the paper.
\end{abstract}

\section{Introduction}

Optical flow estimation, a fundamental problem in computer vision, aims to compute the motion vectors of pixels between consecutive frames in a video sequence. It holds significant and widespread applications in areas such as action recognition~\cite{zhao2020improved,xu2019quadratic}, video interpolation~\cite{huang2022real,wu2022video,liu2020video}, autonomous driving~\cite{yi2023focusflow}, etc.

\begin{figure}[t]
  \centering
  \includegraphics[width=1\linewidth]{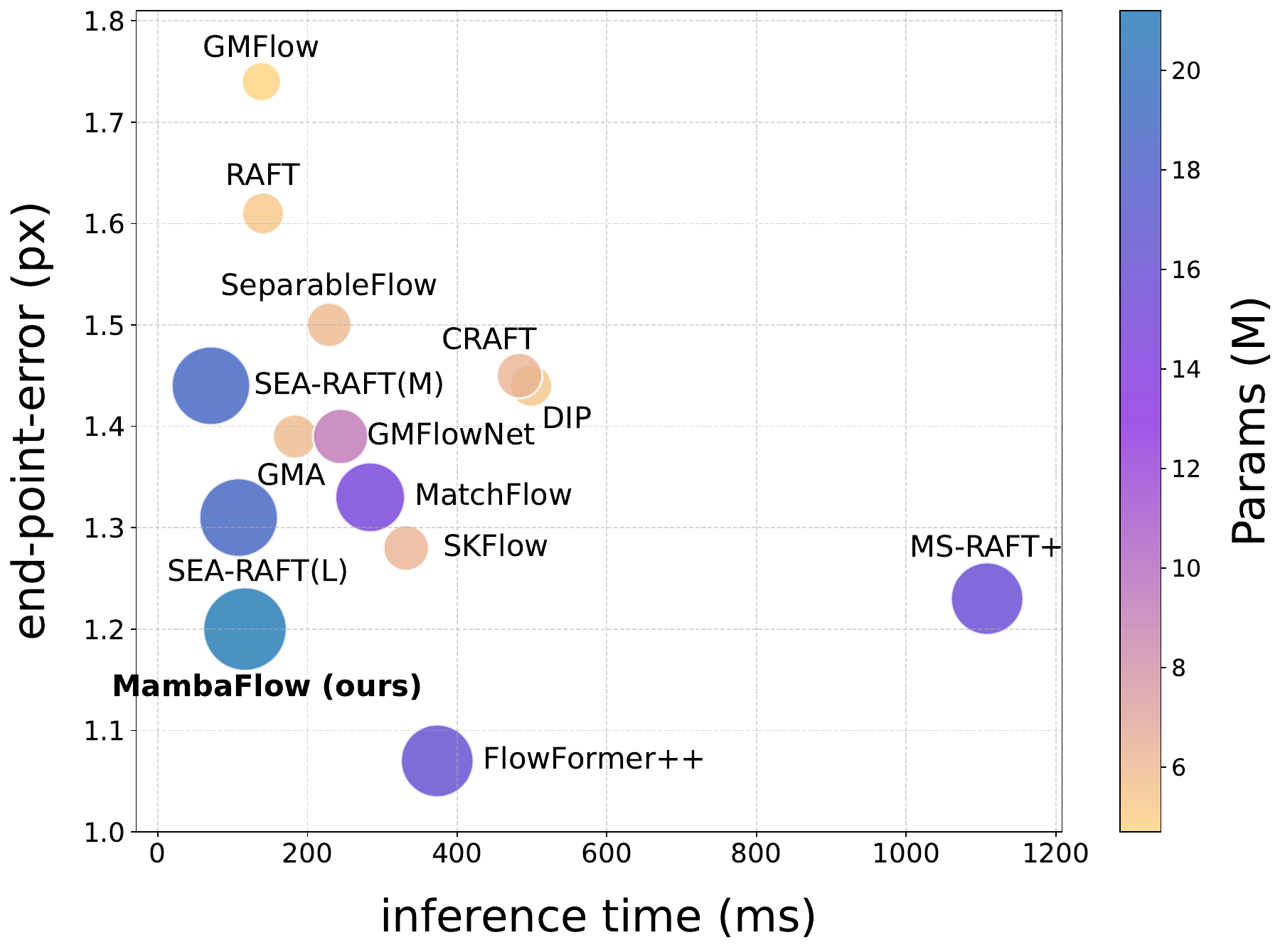}
    \caption{End-point-error on Sintel (clean) vs. inference time (ms) and model params (M). 
    Compared to state-of-the-art models, our proposed model achieves comparable accuracy while significantly reducing inference time, demonstrating an optimal balance between computational efficiency and performance.}
  \label{Fig.1}
\end{figure}

Architectures based on Convolutional Neural Networks (CNNs) have achieved pioneering progress in the field of optical flow. FlowNet~\cite{dosovitskiy2015flownet}, as the first end-to-end CNNs-based flow network, revealed the immense potential of this learning paradigm. However, it faced limitations in handling large displacements and complex motions. Subsequently, PWC-Net~\cite{sun2018pwc} improved estimation accuracy in complex motion scenarios by incorporating pyramid feature extraction and a cost-volume construction, which effectively captured multi-scale local features. Nevertheless, it still encountered challenges in large displacement scenarios. To address these issues, RAFT~\cite{teed2020raft} introduced a recurrent iterative refinement mechanism, which significantly enhanced the accuracy of optical flow estimation. However, such innovation brought a new challenge, which its optimization process relies on numerous iterations, leading to increased parameter counts and longer inference time. Many subsequent variants of the RAFT algorithm have attempted upgrates~\cite{jahedi2024ms,wang2025sea,zheng2022dipdeepinversepatchmatch,sun2022skflow,dong2023rethinkingopticalflowgeometric,jahedi2024ccmr}, such as MS-RAFT and SEA-RAFT, but their ability to handle large displacements and complex motions still requires further improvement.
To tackle the aforementioned issues, the Transformer architecture, known for its powerful global modeling capabilities, has been introduced into optical flow estimation tasks~\cite{huang2022flowformer,shi2023flowformer++,lu2023transflow,sui2022craft}. Most of them leverage attention mechanism to enhance feature representation, which capture long-range dependencies to significantly improve the accuracy. Notably, unlike optical flow algorithms based on RNNs for flow decoding~\cite{teed2020raft,jiang2021learning}, GMFlow~\cite{xu2022gmflow} utilizes the attention mechanism to iteratively decode the final optical flow, resulting in faster flow refinement. However, they encounter the quadratic computational complexity of Transformers both during the training and inference stages~\cite{xu2022gmflow,sui2022craft,huang2022flowformer,shi2023flowformer++,lu2023transflow}, which severely limits their efficiency in practical applications.

Recently, Mamba~\cite{gu2023mamba,gu2021combining} has been widely used in various computer vision tasks, such as image classification~\cite{ge2024mambatsr,yang2024mambamil,yang2024cardiovascular},image segmentation~\cite{liu2024swin,zhou2024efficient,du2024mixed}, object detection~\cite{ma2024fer,ma2024vmambacc,verma2024soar}, image generation~\cite{hu2024zigma,lambrechts2024parallelizing}, video understanding~\cite{jin2024object,li2024videomamba,xiao2024mambatrack}, etc. Those remarkable achievements have demonstrated that Mamba can effectively address the challenges posed by CNNs and Transformers, including the local perception limitations of CNNs and the quadratic computational complexity of Transformers~\cite{xu2024survey,zhang2024survey}. It benefits from a selective state space mechanism, which can efficiently capture global information and handle complex spatiotemporal dependencies with linear complexity~\cite{zhang2024survey}. Specifically, Mamba has been conceptually summarized as being ideally suited for tasks involving long sequences and autoregression~\cite{yu2024mambaout} due to its inherent RNN mechanism within the SSM~\cite{gu2021combining,gu2021efficiently,gu2023mamba}.

Inspired by Mamba's capabilities for long sequence modeling and autoregression with linear complexity, we propose a novel optical flow estimation framework, MambaFlow, designed to address the challenges in CNNs or Transformer-based methods. As demonstrated in Figure~\ref{Fig.1}, our method significantly reduces inference time while maintaining remarkblely high accuracy.The framework introduces two core components: (1) the PolyMamba module, which utilizes Self-Mamba, Cross-Mamba, and Multi-Layer Perceptron (MLP) to globally optimize feature representations with lightweight computation, and (2) the PulseMamba module, which first employs an Attention Guidance Aggregator (AGA) to adaptively integrate motion features using dynamically learned weights—instead of naive concatenation—and then leverages Mamba’s autoregressive capability to iteratively decode and refine optical flow, effectively addressing challenges such as large displacement and occluded regions.

Specifically, we begins with a CNN backbone to extract robust features from the input image pairs similar to traditional practice~\cite{wang2025sea,xu2022gmflow,jiang2021learning}. Subsequently, we leverage the suggested PolyMamba to enhance the extracted features, allowing for the capture of complex global information and long-range dependencies. Building on the enhanced features, we further utilize a global matching mechanism~\cite{xu2022gmflow,zhao2022globalmatchingoverlappingattention} to compute similarities for initial flow. At last, we employ PulseMamba to recurrently refine and yield the final precise flow~\cite{xu2022gmflow}. Our algorithm uses Mamba as the core components in both the encoding and decoding stage, making it an end-to-end optical flow estimation algorithm centered around Mamba. By leveraging Mamba's capabilities in global modeling and autoregressive properties, as well as its computationally efficient characteristics, the proposed approach achieves a good trade-off between accuracy and performance.

Our main contributions can be summarized as follows:

\begin{itemize}
    \item We propose MambaFlow, a novel optical flow estimation architecture built on the Mamba framework both in feature enhancement and flow propagation. To our knowledge, this is the first Mamba-centric architecture designed for end-to-end optical flow estimation. Our method achieves accuracy comparable to state-of-the-art algorithms while significantly reducing training and inference time, which makes it more suitable for real-world application on resource-constrained devices.
    \item We introduce the PolyMamba module, which consists of a Self-Mamba, a Cross-Mamba, and a MLP in sequence, to globally enhance feature representation for both the reference image and its cross features with the matching image. As we will demonstrate, it effectively leverages the Mamba's long sequence modeling capabilities while maintaining a linear computational load.
    \item We design the PulseMamba, which integrates an Attention Guidance Aggregator for adaptive feature fusion and a Mamba-based autoregressive decoder, addressing the computational overload, large displacement and occulusion issues in recurrently refined RNN-like methods. By leveraging the Mamba's powerful autoregressive capability and linear complexity, it offers significant potential to enhance accuracy and speed simultaneously through ablation studies.
\end{itemize}

\section{Related Work}
\subsection{Optical Flow Estimation}

Optical flow estimation was traditionally formulated as an energy minimization task~\cite{brox2004high,horn1981determining,papenberg2006highly}, with early variational methods like Horn and Schunck~\cite{horn1981determining} relying on brightness constancy and spatial smoothness assumptions. Subsequent optimization-based approaches improved robustness through regularization~\cite{bruhn2005lucas}.

In the era of deep learning, following the introduction of FlowNet~\cite{dosovitskiy2015flownet} as the first CNN-based optical flow method, numerous subsequent algorithms such as PWC-Net~\cite{sun2018pwc} and SpyNet~\cite{ranjan2017optical}, adopted a coarse-to-fine strategy for optical flow estimation. However, these approaches struggled with fast motions and small displacements due to inaccurate coarse-level guidance. RAFT~\cite{teed2020raft} addressed this via iterative full-field refinement, inspiring variants such as GMA~\cite{jiang2021learning}, SeparableFlow~\cite{zhang2021separable}, and MS-RAFT~\cite{jahedi2024ms}, which introduced modules to better handle occlusion and motion ambiguity. Despite their accuracy, these CNN-based methods require heavy iterative computation and struggle with large displacements and complex motion, limiting their efficiency on resource-constrained devices.

Transformers have emerged as powerful alternatives due to their ability to model global dependencies via self-attention. Models like FlowFormer~\cite{huang2022flowformer}, CRAFT~\cite{sui2022craft}, and GAFlow~\cite{luo2023gaflow} leveraged Transformer architectures to enhance feature expressiveness and global matching. GMFlow~\cite{xu2022gmflow} and TransFlow~\cite{lu2023transflow} further improved efficiency by reformulating optical flow as a one-shot global matching problem. However, the quadratic complexity of attention mechanisms results in high computational and memory costs, limiting deployment on lightweight systems.

\subsection{State Space Models}

State space models~\cite{gu2022efficientlymodelinglongsequences}, which originated from control theory, have been effectively integrated with deep learning to capture long-range dependencies. 
Mamba~\cite{gu2023mamba} introduced an input adaptation mechanism to enhance the state space model, providing faster inference speed, higher throughput and overall indicators compared to the Transformers of the same size, especially in natural language. Related work on Mamba~\cite{wu2025cross} has shown that it has comparable performance to the Transformer-based methods while significantly reducing computational complexity. This superiority led to further exploration like Vision Mamba~\cite{zhu2024vision} and Vmamba~\cite{liu2024vmambavisualstatespace}, which applied SSMs to visual tasks such as image classification~\cite{liu2024vmambavisualstatespace,zhu2024vision}, video understanding~\cite{li2025videomamba,wang2023selective}, biomedical image segmentation~\cite{ma2024u,xing2024segmamba} and others~\cite{hu2025zigma,islam2023efficient,nguyen2022s4nd,qin2024mambavc,zha2024lcm}. 
However, their applicability to optical flow estimation has remained unexplored until recently. Lin et al.~\cite{lin2024flowmamba} pioneered the use of Mamba in the scene flow estimation task, proposing an SSM-based iterative update module to replace the GRU module as a decoder. Nevertheless, it was specifically designed for scene flow with point cloud feature and is not suitable for the generally used backbone optical flow methods.

Inspired by the Mamba's strength of long sequence modeling and autoregression with linear complexity, we propose a Mamba-centric architecture for optical flow estimation. Our goal is to design an end-to-end solution that utilizes Mamba for feature enhancement and flow optimization. These Mamba-centric designs guarantee the accuracy of our method while maintaining its high speed in training and inference, providing valuable insights for resource-constrained devices.

\begin{figure*}[t] 
    \centering
    \includegraphics[width=\textwidth]{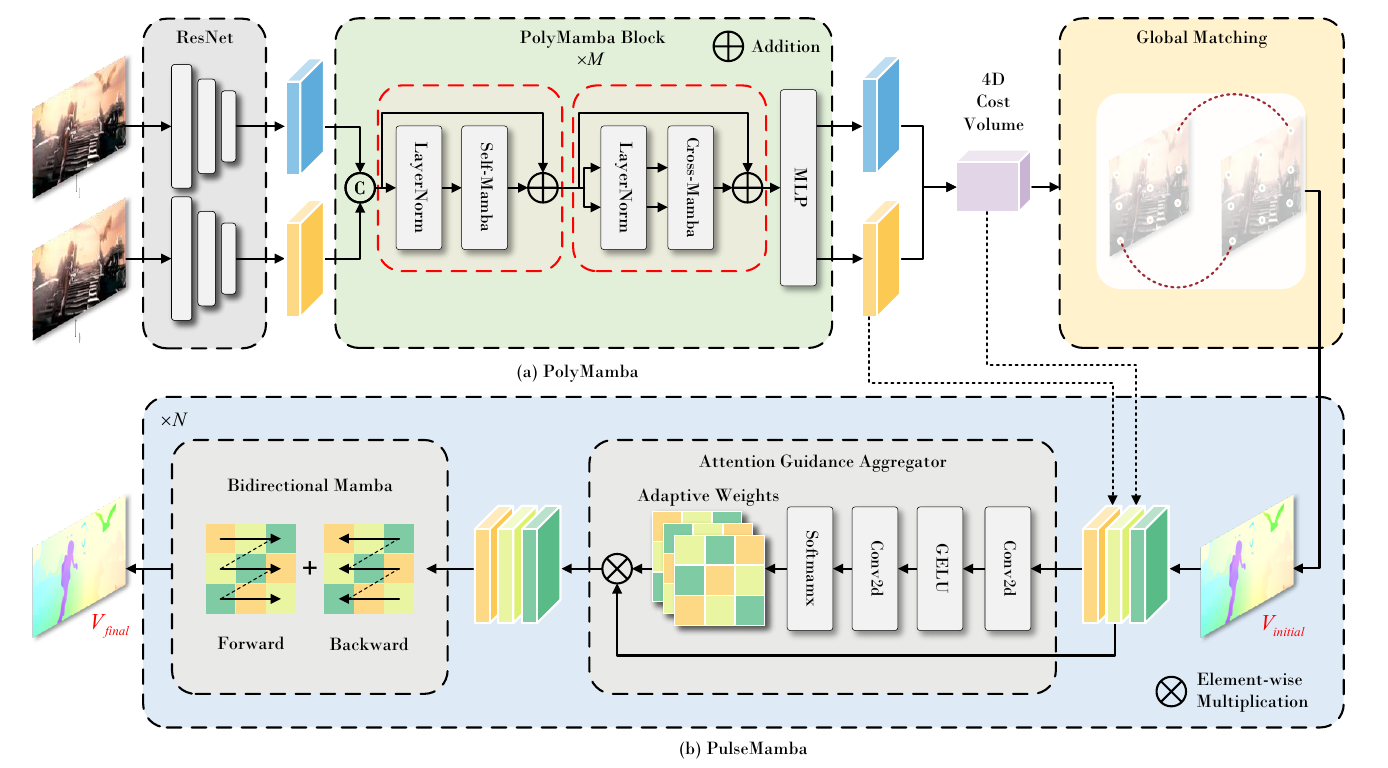} 
    \caption{The overall network architecture of our MambaFlow.}  
    \label{Fig.2} 
\end{figure*}

\section{Method}
\subsection{Preliminaries}

SSMs are inspired by linear systems, mapping an input sequence \(x_t \in \mathbb{R}\) to an output \(y_t \in \mathbb{R}\) through a hidden state \(h_t \in \mathbb{R}^N\). The model can be formulated as linear ordinary differential equations (ODEs):
\begin{equation}
h_{t} = \overline{\mathbf{A}} h_{t-1} + \overline{\mathbf{B}} x_{t},
\label{Eq.1}
\end{equation}
\begin{equation}
y_{t} ={\mathbf{C}} h_{t},
\label{Eq.2}
\end{equation}
where $\overline{\mathbf{A}}$ and $\overline{\mathbf{B}}$ represent the discretized parameters derived from the given parameters \(\mathbf{A} \in \mathbb{R}^{N \times N}\), \(\mathbf{B} \in \mathbb{R}^{N \times 1}\), and the time scale parameter $\boldsymbol{\Delta}$:
\begin{equation}
\overline{\mathbf{A}}=\exp (\boldsymbol{\Delta} \mathbf{A}),
\label{Eq.3}
\end{equation}
\begin{equation}
\overline{\mathbf{B}}=(\boldsymbol{\Delta} \mathbf{A})^{-1}(\exp (\boldsymbol{\Delta} \mathbf{A})-\mathbf{I}) \cdot \boldsymbol{\Delta} \mathbf{B}.
\label{Eq.4}
\end{equation}

Additionally, \(\mathbf{C} \in \mathbb{R}^{1 \times N}\) serves as the projection matrix mapping the hidden state \(h_t\) to the output \(y_t\).

Finally, the models generate the output using a global convolution operation, defined as:
\begin{equation}
\overline{\mathbf{K}} = (\mathbf{C} \overline{\mathbf{B}}, \mathbf{C} \overline{\mathbf{A}} \overline{\mathbf{B}}, \ldots, \mathbf{C} \overline{\mathbf{A}}^{M-1} \overline{\mathbf{B}}),
\label{Eq.5}
\end{equation}
\begin{equation}
y = x \ast \mathbf{\overline{K}},
\label{Eq.6}
\end{equation}
where $M$ represents the length of the input sequence $x$, and $\mathbf{\overline{K}} \in \mathbb{R}^M$ denotes a structured convolutional kernel.

The recently proposed advanced state-space model, Mamba, enhances \(\mathbf{B}\), \(\mathbf{C}\), and \(\boldsymbol{\Delta}\) by making them input-dependent, thereby enabling dynamic feature representations. Specifically, these matrices are defined as follows:
\begin{equation}
    \mathbf{B}_i = S_B(x_i), 
\label{Eq.7}
\end{equation}
\begin{equation}
    \mathbf{C}_i = S_C(x_i), 
\label{Eq.8}
\end{equation}
\begin{equation}
    \boldsymbol{\Delta}_i = \text{Softplus}(S_\Delta(x_i)),
\label{Eq.9}
\end{equation}
where $S_B$, $S_C$, and $S_\Delta$ are linear projection functions, and $i$ denotes the index of the $i$-th element in the sequence. 

Simultaneously, the parallel scan algorithm enables Mamba to leverage the same parallel processing advantages as described in \eqref{Eq.5}, thereby facilitating efficient training.

\subsection{Overall Architecture}

Optical flow can be intuitively understood as a task aimed at finding pixel-wise correspondences between consecutive images. As shown in Figure~\ref{Fig.2}, our proposed framework, MambaFlow, is centered around two novel modules: the feature enhancement module PolyMamba and the flow refinement module PulseMamba.

Given two input frames \( \boldsymbol{I}_1 \) and \( \boldsymbol{I}_2 \), we first extract dense feature maps \( \boldsymbol{F}_1, \boldsymbol{F}_2 \in \mathbb{R}^{H \times W \times D} \) using a shared-weight convolutional encoder, where \(H\), \(W\), and \(D\) denote the height, width, and feature dimension of the features. These features are then processed by the PolyMamba module, which stacks Self-Mamba and Cross-Mamba blocks to capture both intra-frame and inter-frame dependencies. To further enhance representation capacity, a lightweight MLP layer is appended after each block group for local-global feature fusion. The resulting features \( \boldsymbol{\mathit{F}}_q \) and \( \boldsymbol{\mathit{F}}_v \in \mathbb{R}^{H \times W \times D} \) are then used for global matching to estimate the initial flow.

The initial prediction \( \mathbf{V}_{\text{initial}} \) is then refined by the PulseMamba module, which incorporates cost volume and motion features into a stack of Mamba layers for iterative flow refinement. This module significantly improves performance in occluded or textureless regions, enhancing both the accuracy and robustness of optical flow estimation.

\subsection{Self-Mamba}
The input features $\boldsymbol{\mathit{F}}_1, \boldsymbol{\mathit{F}}_2 \in \mathbb{R}^{H \times W \times D}$ are enriched with learnable positional embeddings to retain spatial positional information. Subsequently, $\boldsymbol{\mathit{F}}_1$ and $\boldsymbol{\mathit{F}}_2$ are processed through a weight-sharing Self-Mamba block, yielding enhanced representations denoted as $\boldsymbol{\mathit{F}}_1^{\text{self}}$ and $\boldsymbol{\mathit{F}}_2^{\text{self}}$, respectively.

The processing within each Self-Mamba block can be summarized as a bidirectional operation, where the forward and backward hidden states are computed and combined to enrich the feature representation. The overall process is expressed as:

\begin{equation}
\begin{aligned}
    \boldsymbol{\mathit{F}}_i^{\text{self}}
        &= \text{Self-Mamba}(\boldsymbol{\mathit{F}}_i) \\
        &= \text{Self-Forward}(\boldsymbol{\mathit{F}}_i) 
         + \text{Self-Backward}(\boldsymbol{\mathit{F}}_i) \\
        &\quad i \in \{1, 2\}.
\end{aligned}
\label{Eq.10}
\end{equation}

Here, the \(\text{Self-Mamba}\) function represents the state space model used to compute the forward and backward hidden states, with the sequence $\boldsymbol{\mathit{F}_i}$ being reversed to distinguish between the forward and backward directions.

\subsection{Cross-Mamba}

While the Self-Mamba architecture is effective in modeling long-range dependencies within a single sequence, it lacks the capacity to capture inter-sequence relationships—an ability that is crucial in optical flow estimation, where information from two correlated feature maps must be jointly reasoned. To address this limitation, we introduce a cross-sequence interaction module based on selective state-space modeling, enabling rich bidirectional information exchange between paired features.

Given the enhanced features
\(
\boldsymbol{\mathit{F}}_1^{\text{self}},
\boldsymbol{\mathit{F}}_2^{\text{self}}
\),
the main branch applies a depth-wise causal convolution and activation:
\begin{equation}
    \boldsymbol{x}_{\mathrm{conv}} 
    = \operatorname{SiLU} \left( \operatorname{DWConv} ( \boldsymbol{\mathit{F}}_1^{\text{self}} ) \right),
    \label{Eq.11}
\end{equation}
while the modulation branch linearly projects the second stream:
\begin{equation}
    \boldsymbol{x}_{\mathrm{mod}} = \operatorname{Proj} ( \boldsymbol{\mathit{F}}_2^{\text{self}} ).
    \label{Eq.12}
\end{equation}

Their concatenation is projected into three parameter groups:
\begin{equation}
    [\, \boldsymbol{\Delta},\; \mathbf{B},\; \mathbf{C} \,] = 
    \operatorname{Linear} ( [\boldsymbol{x}_{\mathrm{conv}};\; \boldsymbol{x}_{\mathrm{mod}}] ),
    \label{Eq.13}
\end{equation}

The hidden state $\boldsymbol{y}$ is then updated through a SelectiveScan kernel, which dynamically integrates temporal and spatial information using parameterized state-space operations.

A final linear projection yields the cross-attended representation:
\begin{equation}
    \boldsymbol{\mathit{F}}^{\text{cross}} = \operatorname{OutProj} ( \boldsymbol{y} ).
    \label{Eq.14}
\end{equation}

To model temporal symmetry, Cross-Mamba is applied in both forward and backward directions—just as in the Self-Mamba block—and the outputs are summed:

\begin{equation}
\begin{aligned}
    &\text{Cross-Mamba} \left( 
        \boldsymbol{\mathit{F}}_1^{\text{self}},\;
        \boldsymbol{\mathit{F}}_2^{\text{self}}
    \right) \\
    &=\; \text{Forward} (\,\cdot\,) + \text{Backward} (\,\cdot\,).
\end{aligned}
\label{Eq.15}
\end{equation}

\subsection{Global Matching}

After extracting the enhanced features $\boldsymbol{\mathit{F}}_{q}$ and $\boldsymbol{\mathit{F}}_{v}$ from our PolyMamba module, we construct a 4D cost volume of size $H \times W \times H \times W$ by computing the dot-product similarity between all pixel pairs from both features. The constructed 4D cost volume can be interpreted as a collection of $H \times W$ 2D cost maps, where each map evaluates the feature similarity between a single pixel in $\boldsymbol{\mathit{F}}_{q}$ and all pixels in $\boldsymbol{\mathit{F}}_{v}$. To obtain a matching distribution, a softmax operation is applied across the last two dimensions of the cost volume. The construction process is expressed as:
\begin{equation}
\mathbf{M} = \text{softmax}(\mathbf{\frac{\boldsymbol{\mathit{F}}_{\mathit{q}} \, \boldsymbol{\mathit{F}}_{\mathit{v}}^\top}{\sqrt{D}}}).
\label{Eq.16}
\end{equation}

where $D$ is the feature dimension. To estimate the optical flow, the correspondence is obtained by taking a weighted average of the 2D pixel grid coordinates $\mathbf{G} \in \mathbb{R}^{H \times W \times 2}$ in the target image with the matching distribution $M$. Finally, the optical flow $\mathbf{V}_{\text{initial}}$ is calculated as the difference between the corresponding pixel coordinates:
\begin{equation}
\mathbf{V}_{\text{initial}} = \mathbf{M} \mathbf{G} - \mathbf{G}.
\label{Eq.17}
\end{equation}

The global matching method effectively improves the accuracy and stability of optical flow estimation by capturing global contextual information.

\begin{table*}[ht]
\centering
\setlength{\tabcolsep}{8pt}  
\begin{tabular}{llrccccc}
\toprule
\multirow{2}{*}{\textbf{Extra Data}} 
& \multicolumn{2}{c}{\multirow{2}{*}{\textbf{Method}}}
& \multicolumn{2}{c}{\textbf{Sintel}} 
& \textbf{KITTI} 
& \multicolumn{2}{c}{\textbf{Inference Cost}} \\
\cmidrule(lr){4-5} \cmidrule(lr){6-6} \cmidrule(lr){7-8}
& & & Clean$\downarrow$ & Final$\downarrow$ & Fl-all$\downarrow$ & Param & Time \\
\midrule
  & RAFT & ECCV'20 & 1.61 & 2.86  & 5.10  & 5.3M & 140.7ms \\
  & SeparableFlow & ICCV'21 & 1.50 & 2.67  & \underline{4.64}  & 6.0M & 229.1ms \\
  & GMA & ICCV'21 & 1.39 & 2.47  & 5.15  & 5.9M & 183.3ms \\
  & DIP & CVPR'22 & 1.44 & 2.83  & 4.21  & 5.4M & 498.9ms \\
  & GMFlowNet & CVPR'22 & 1.39 & 2.65  & 4.79  & 9.3M & 244.3ms \\
  & CRAFT & CVPR'22 & 1.45 & \underline{2.42}  & 4.79  & 6.3M & 483.4ms \\
  & GMFlow & CVPR'22 & 1.74 & 2.90  & 9.32  & 4.7M & \underline{138.5ms} \\
  & SKFlow & NeurIPS'22 & 1.28 & \textbf{2.23}  & 4.85  & 6.3M & 331.9ms \\
  & MatchFlow & CVPR'23 & 1.33 & 2.64 & 4.72  & 14.8M & 283.8ms \\
  & MS-RAFT+ & IJCV'24 & \underline{1.23} & 2.68 & \textbf{4.15} & 16.0M & 1108.2ms \\
  & \textbf{MambaFlow (ours)} &  & \textbf{1.20} & 2.62 & 7.34  & 20.5M & \textbf{116.5ms} \\
\midrule
CroCo-Pretrain & CroCoFlow & ICCV'23 & \underline{1.09} & 2.44  & \underline{3.64}  & 437.4M & 6422.0ms \\
DDVM-Pretrain & DDVM & NeurIPS'23 & 1.75 & 2.48  & \textbf{3.26}  & -- & -- \\
YouTube-VOS & FlowFormer++ & CVPR'23 & \textbf{1.07} & \textbf{1.94}  & 4.52  & 16.1M & 373.4ms \\
VIPER & CCMR+ & WACV'24 & \textbf{1.07} & \underline{2.10}  & 3.86  & 11.5M & OOM \\
TartanAir & SEA-RAFT(M) & ECCV'24 & 1.44 & 2.86  & 4.64  & 18.8M & \textbf{71.0ms} \\
TartanAir & SEA-RAFT(L) & ECCV'24 & 1.31 & 2.60  & 4.30  & 18.8M & \underline{108.0ms} \\
\midrule
\end{tabular}
\caption{
Quantitative results on Sintel and KITTI benchmarks. All methods are trained on the unified C+T+S+K+H datasets as the baseline. The "Extra Data" column indicates additional pretraining datasets. Evaluation is conducted with batch size 1, input resolution $540 \times 960$, and an RTX 3090 GPU.
}
\label{Table.1}
\end{table*}

\begin{figure*}[t] 
    \centering
    \includegraphics[width=\linewidth]{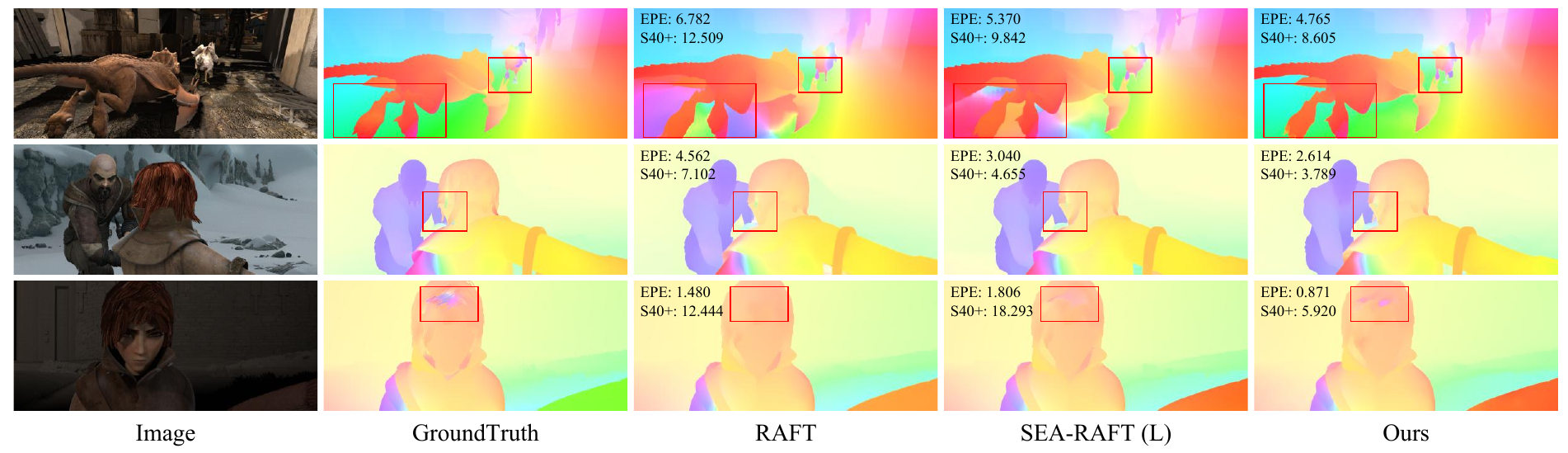} 
    \caption{Visual results on Sintel test set.}
    \label{Fig.3} 
\end{figure*}

\subsection{PulseMamba}

To effectively address occlusions and ambiguous boundary regions in optical flow estimation, we propose PulseMamba, an autoregressive refinement module featuring an Attention Guidance Aggregator (AGA) for adaptive feature integration. While conventional feature concatenation equally weights all spatial information, it often neglects dynamically relevant regions. To address this limitation, AGA adaptively assigns spatial attention weights to each feature, enabling context-aware flow estimation.

Given initial optical flow \(\mathbf{V}_{\text{initial}}\), source image features \(\boldsymbol{F}_q\), and a 4D correlation cost volume \(\mathbf{C}\), PulseMamba first extracts motion features:
\begin{equation}
    \boldsymbol{M} = \text{MotionEncoder}(\mathbf{V}_{\text{initial}}, \mathbf{C}).
    \label{Eq.18}
\end{equation}

AGA adaptively integrates motion features \(\boldsymbol{M}\), context features \(\boldsymbol{F}_q\), and hidden-state features \(\boldsymbol{h}\) into a contextually informative representation. Specifically, AGA first aligns these input features to a unified embedding dimension through individual convolutional projections. 

Then, concatenated aligned features are processed via convolutional and GELU layers to produce spatial attention maps \(\boldsymbol{A}\):

\begin{equation}
[\boldsymbol{M};\, \boldsymbol{F}_q;\, \boldsymbol{h}]
\xrightarrow{\mathrm{Conv}_1}
\xrightarrow{\mathrm{GELU}}
\xrightarrow{\mathrm{Conv}_2}
\xrightarrow{\mathrm{Softmax}}
\boldsymbol{A.}
\label{Eq.attention}
\end{equation}

Finally, the aggregated feature \(\boldsymbol{x}_{\text{AGA}}\) is obtained by a weighted sum:
\begin{equation}
    \boldsymbol{x}_{\text{AGA}} = \sum_{f \in \{M, F_q, h\}} \boldsymbol{A}_{f} \odot \boldsymbol{f},
    \label{Eq.AGA_final}
\end{equation}
where \(\boldsymbol{A}_{f}\) represents spatial attention weights corresponding to each feature type. 

The aggregated feature \(\boldsymbol{x}_{\text{AGA}}\) is then iteratively refined by PulseMamba using a Mamba layer to update the hidden-state representation:
\begin{equation}
    \boldsymbol{h}^{(i)}=\text{Mamba}(\boldsymbol{x}_{\text{AGA}}^{(i-1)}).
    \label{Eq.Mamba}
\end{equation}

Subsequently, the flow increment \(\Delta\mathbf{V}^{(i)}\) is predicted, and the optical flow is iteratively updated:

\begin{equation}
\begin{aligned}
    \Delta \mathbf{V}^{(i)} &= \text{FlowHead}(\boldsymbol{h}^{(i)}) \\
    \mathbf{V}^{(i)} &= \mathbf{V}^{(i-1)} + \Delta\mathbf{V}^{(i)}.
\end{aligned}
\label{Eq.update}
\end{equation}

After \(N\) iterations, PulseMamba produces the final refined optical flow:
\begin{equation}
    \mathbf{V}_{\text{final}}=\mathbf{V}^{(N)}.
    \label{Eq.final}
\end{equation}

Through dynamic attention-guided feature integration and autoregressive refinement, PulseMamba effectively enhances the robustness and precision of optical flow estimation, especially under challenging scenarios.

\section{Experiments}
\paragraph{Datasets.} 
Following the previous methodology, we first pre-train on the FlyingChairs (Chairs)\cite{dosovitskiy2015flownet} and FlyingThings3D (Things)\cite{mayer2016large} datasets. For fine-tuning, we adopt a mixed training set consisting of KITTI~\cite{menze2015object}, HD1K~\cite{kondermann2016hci}, FlyingThings3D, and Sintel~\cite{butler2012naturalistic} training sets. Finally, we further fine-tune the model on the KITTI training set and report the performance on the respective online benchmarks.

\paragraph{Metrics.} 
In our experiments, we employed evaluation metrics commonly used in optical flow estimation. Firstly, we utilized the end-point-error (EPE) to measure the average Euclidean distance between the predicted and ground truth optical flow vectors. For the KITTI dataset, we adopted the F1-all metric, which represents the percentage of pixels with an error exceeding 3 pixels or a relative error greater than 5\%. To gain deeper insights into the model's performance under large motion magnitudes, we defined $s_{40+}$ as the EPE for pixels where the ground truth flow magnitude exceeds 40 pixels.

\paragraph{Implementation Details.}
Our model is constructed using the PyTorch library and trained on 8 NVIDIA GeForce RTX 3090 GPUs. Following the training strategy from previous works~\cite{xu2022gmflow,teed2020raft}, we use the AdamW optimizer. The model is first pre-trained on the Chairs dataset for 100k iterations with a batch size of 16 and a learning rate of $4 \times 10^{-4}$. Then, it is fine-tuned on the Things dataset for 200k iterations with a batch size of 8 and a learning rate of $2 \times 10^{-4}$. During this process, we select the best-performing model on the Things dataset and fine-tune it for an additional 800k iterations to validate the ablation study. Subsequently, the model fine-tuned for 200k iterations on the Things dataset is used as the initial model and further trained on the Sintel dataset, keeping the same hyperparameters as those used for the Things dataset, including a batch size of 8 and a learning rate of $2 \times 10^{-4}$. Finally, we continue training on the KITTI-2015 training set for 200k iterations with the batch size and learning rate unchanged.

\subsection{Comparison with State-of-the-Arts}

We compared the performance of different optical flow algorithms on the Sintel and KITTI datasets, and the experimental results are shown in Table~\ref{Table.1}. 

Under the unified training setting (C+T+S+K+H), we first compare MambaFlow with methods that do not use additional pretraining data. MambaFlow achieves an EPE of 1.20 on Sintel Clean and 2.62 on Sintel Final, outperforming MS-RAFT+ in both accuracy and inference speed—being 9.51$\times$ faster (116.5ms vs 1108.2ms), and achieves leading efficiency among non-pretrained methods.

Among the methods that utilize extra pretraining data, SEA-RAFT is recognized as one of the fastest baselines. Our model achieves a similar inference speed to SEA-RAFT(L). Despite the use of additional TartanAir data in SEA-RAFT(L), MambaFlow achieves a lower EPE on Sintel Clean, with an 8.4\% reduction compared to SEA-RAFT(L). For other methods that utilize large-scale extra pretraining data, such as FlowFormer++ (YouTube-VOS\cite{xu2018youtube}), CroCoFlow\cite{weinzaepfel2023croco} (CroCo-Pretrain), and others, MambaFlow achieves comparable accuracy while maintaining significantly faster inference.

Figure~\ref{Fig.3} presents a qualitative analysis of optical flow estimation, comparing MambaFlow with other state-of-the-art approaches on the Sintel test set.

Overall, these results demonstrate that MambaFlow achieves a favorable trade-off between accuracy and efficiency, and delivers performance comparable to state-of-the-art methods, regardless of whether additional pretraining data is utilized.

\begin{table}[t]
\centering
\setlength{\tabcolsep}{7pt}
\renewcommand{\arraystretch}{1.0}
\begin{tabular}{lcccc}
\toprule
\multirow{2}{*}{\textbf{Setup}} 
& \multicolumn{1}{c}{\textbf{Things (val)}} 
& \multicolumn{2}{c}{\textbf{Sintel (train)}} 
& \textbf{Param} \\
\cmidrule(lr){2-2} \cmidrule(lr){3-4}
& clean 
& clean & final 
& (M) \\
\midrule
full       & \textbf{2.22} & \textbf{1.15} & \textbf{3.05} & 20.5 \\
w/o self   & 2.40 & 1.24 & 3.12 & 18.2 \\
w/o cross  & 3.84 & 2.00 & 4.61 & 14.7 \\
w/o MLP    & 2.46 & 1.25 & 3.16 & 17.5 \\
w/o pos    & 2.29 & 1.18 & 3.11 & 20.5 \\
\bottomrule
\end{tabular}
\caption{Ablations on the Components of PolyMamba}
\label{Table.2}
\end{table}

\begin{table}[t]
\centering
\setlength{\tabcolsep}{5pt}
\renewcommand{\arraystretch}{1}
\begin{tabular}{cccccc}
\toprule
\multirow{2}{*}{\textbf{Blocks}} 
& \multicolumn{2}{c}{\textbf{Sintel (train)}} 
& \multicolumn{2}{c}{\textbf{KITTI (train)}} 
& \multirow{2}{*}{\textbf{Param (M)}} \\
\cmidrule(lr){2-3} \cmidrule(lr){4-5}
& Clean$\downarrow$ & Final$\downarrow$ & EPE$\downarrow$ & F1-all$\downarrow$ & \\
\midrule
4  & 1.20 & 3.21 & 7.54 & 22.28 & 16.5 \\
6  & 1.16 & 3.17 & 7.27  & 20.81 & 18.5 \\
\textbf{8}  & \textbf{1.15} & \textbf{3.05} & \textbf{6.76}  & \textbf{20.50} & \textbf{20.5} \\
10 & 1.14 & 3.11 & 6.71 & 20.61 & 22.5 \\
12 & 1.15 & 2.96 & 6.45 & 18.90 & 24.5  \\
\bottomrule
\end{tabular}
\caption{Ablations on the Number of PolyMamba Blocks}
\label{Table.3}
\end{table}

\begin{table}[t]
\centering
\setlength{\tabcolsep}{5pt}
\begin{tabular}{ccccc}
\toprule
\multirow{2}{*}{\makecell{\textbf{PulseMamba}\\\textbf{iterations}}}
& \multirow{2}{*}{\makecell{\textbf{use}\\\textbf{AGA?}}} & \multicolumn{3}{c}{\textbf{Sintel (train, clean)}} \\
\cmidrule(lr){3-5}
& & all & matched & unmatched \\
\midrule
0                  & --     & 1.96 & 0.68 & 15.75   \\
\midrule
\multirow{2}{*}{1} & \ding{51}    & 0.62 & 0.38 & 3.22    \\
                   & \ding{55}     & 0.72 & 0.45 & 3.58    \\
\midrule
\multirow{2}{*}{2} & \ding{51}    & 0.56 & 0.34 & 2.99    \\
                   & \ding{55}     & 0.60 & 0.36 & 3.16    \\
\midrule
\multirow{2}{*}{3} & \ding{51}    & 0.55 & 0.33 & 2.93    \\
                   & \ding{55}     & 0.57 & 0.35 & 2.91    \\
\bottomrule
\end{tabular}
\caption{Ablations on PulseMamba Designs}
\label{Table.4}
\end{table}

\subsection{Ablation Analysis}
\paragraph{Components of PolyMamba.}
We ablate different PolyMamba components in Table~\ref{Table.2}. Among them, Cross-Mamba contributes the most as it models the mutual relationship between two features, a relationship that is otherwise missing in the features extracted from the convolutional backbone.  Self-Mamba aggregates contextual cues within the same feature, resulting in an additional performance improvement, and positional information introduces location dependence into the matching process, helping to mitigate ambiguities present in feature-similarity-based matching alone. Removing the MLP reduces the number of parameters, but also results in a slight decrease in performance.

\paragraph{Number of PolyMamba blocks.} 
We conduct an ablation study on the number of PolyMamba blocks, as shown in Table~\ref{Table.3}. The experiment compares the impact of varying the number of PolyMamba blocks on overall performance. Among all the tested configurations, the PolyMamba with 8 blocks achieves a good balance between accuracy and parameters.

\paragraph{Ablation Study on PulseMamba and AGA.}

Table~\ref{Table.4} shows that applying PulseMamba iterations brings substantial gains, particularly for unmatched pixels. A single iteration reduces overall EPE from 1.96 to 0.62, confirming the effectiveness of the design, and additional iterations further reduce the overall error. AGA consistently outperforms simple concatenation across all configurations; for example, with two iterations, the overall EPE decreases from 0.60 (concatenation) to 0.56 (AGA), demonstrating the benefit of adaptive feature fusion. Based on these results, we adopt two PulseMamba iterations with AGA as the default in our final model.

\section{Conclusion}
As Mamba continues to gain significant progresses in the field of computer vision, this paper explores for the first time to leverage it to estimate optical flow in an end-to-end paradigm. We suggest a Mamba-centric optical flow method, MambaFlow, which achieves a good balance between accuracy and performance compared to state-of-the-art methods. We attribute this to the Mamba's capabilities in global modeling and autoregression, as well as its linear complexity, which offers greater potential for real-world deployment on resource-constrained devices, and provides valuable insights that may influence ongoing and future developments in this rapidly evolving field of optical flow.

\paragraph{Limitations.}

While MambaFlow achieves results comparable to some state-of-the-art methods, there is still room for improvement, as indicated by the KITTI results in Table~\ref{Table.1}. The model’s generalization may be limited when there is a substantial gap between training and test data, and its performance under challenging conditions such as motion blur, fog, or high-resolution images remains unexplored. We plan to pretrain on large-scale datasets like TartanAir~\cite{wangtartanair} and VIPER~\cite{richter2017playing} to address these issues. Additionally, with the emergence of many Mamba variants~\cite{zhang2024survey}, we see potential to further refine the architecture and core Mamba-based components for improved accuracy and real-world applicability.

\bibliography{aaai2026}
\end{document}